\documentclass[12pt]{article}

\usepackage{amsmath,amssymb,amsthm,mathtools}
\usepackage{setspace,enumitem,soul,url,xcolor,hyperref,bm,stackrel}
\usepackage{tikz-cd}

\usepackage[T1]{fontenc}
\usepackage{tgbonum} \fontfamily{qbk} 

\begin{document}

\title{Graph sampling for node embedding}
\author{\emph{Li-Chun Zhang}}
\date{}
\maketitle

\begin{abstract}
Node embedding is a central topic in graph representation learning. Computational efficiency and scalability can be challenging to any method that requires full-graph operations. We propose sampling approaches to node embedding, with or without explicit modelling of the feature vector, which aim to extract useful information from both the eigenvectors related to the graph Laplacien and the given values associated with the graph.
\end{abstract}


\section{Introduction} 

Valued graph is a ubiquitous form of data. Graph representation learning can be used for node classification, edge prediction, community detection and various other purposes; see e.g. Hamilton (2020). A key to all the most successful approaches is to utilise the links between nodes, rather than simply treating the nodes as statistically independent units. In this context, node embedding either refers to the generation of a low-dimension feature vector for each node in the given graph or the generated vectors as inputs to downstream tasks, which aims to capture the information about a node's position and neighbourhood structure in the graph as well as the associated values. 

Node embedding can be obtained by so-called traditional methods, which do not require an explicit model of the features being targeted, such as based on factorisation or random walks; see e.g. Khoshraftar and An (2022), Hamilton (2020). More recently, the modelling approach has become very popular, where embeddings are typically obtained using graph neural networks (GNN). See e.g. Zhou et al. (2020) for a review of GNN methods and applications.

\begin{figure}[ht]
\centering
\includegraphics[scale=0.6]{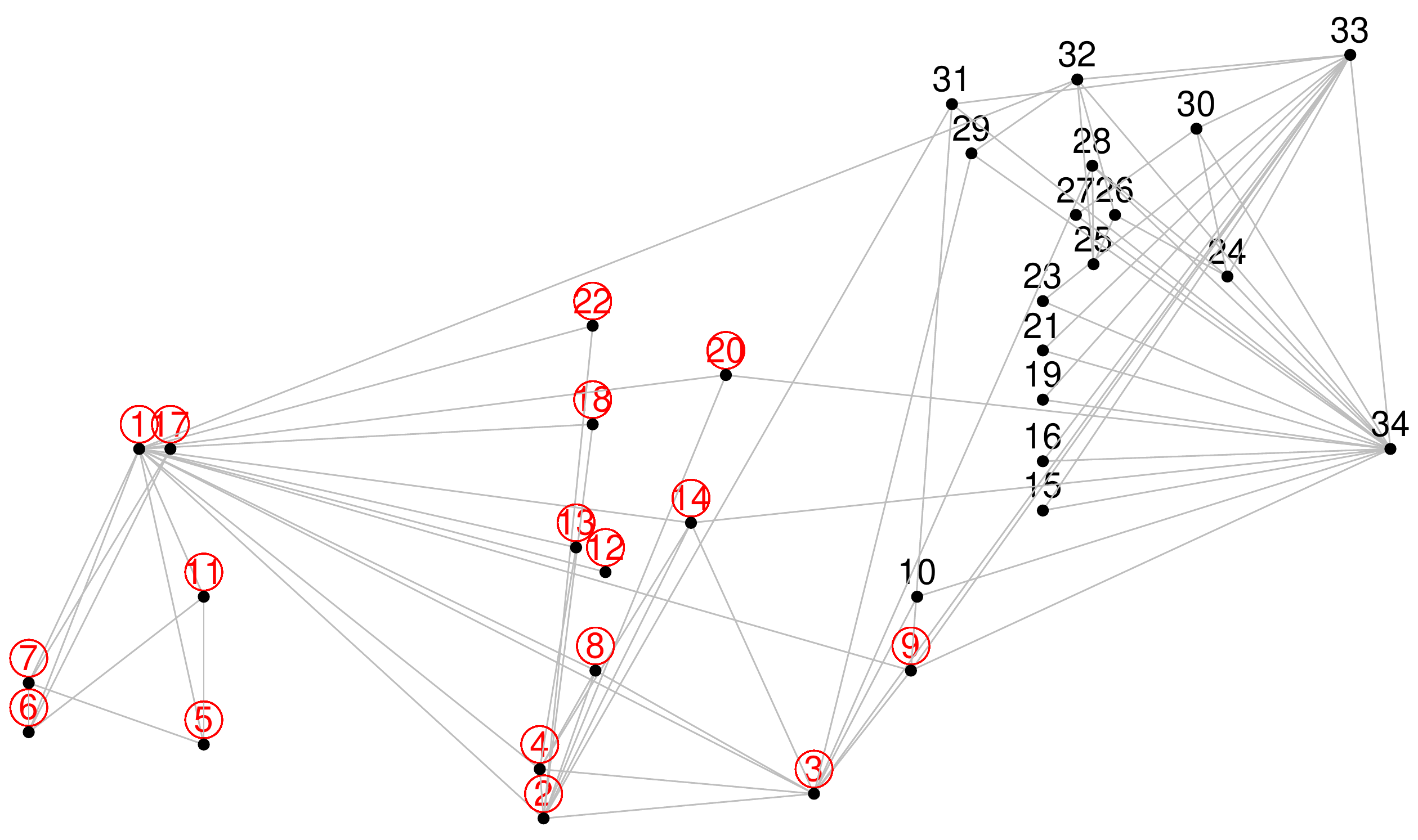} 
\caption{Members split (encircled or not) of Zachary's karate club} \label{fig:ZKC}
\end{figure}

Figure \ref{fig:ZKC} shows the well-known valued graph of Zachary's karate club (ZKC), where edges represent social relationships between the 34 members (i.e. nodes) which are indicative of the members split (Zazhary, 1977). As an example of node embedding discussed in the literature, every node can be assigned one of four classes by modularity-based clustering (Brandes et al., 2008). Kipf and Welling (2017) show that this community structure can be closely captured by 2-vector node embeddings, using a 3-layer graph convolutional networks (GCN), and these embeddings are comparable to those obtained by DeepWalk (Perozzi et al., 2014) which does not require explicit modelling.

Meanwhile, the $x$-coordinates of the nodes in Figure \ref{fig:ZKC} are actually given by the eigenvector corresponding to the smallest non-zero eigenvalue of the normalised Laplacian of this graph, which can be attributed to the traditional factorisation-based Laplacian eigenmaps technique of Belkin and Niyogi (2001). This illustrates clearly how embedding can facilitate node classification, since perfect classification of the members split can be obtained using these scalar $x_i$ as the node embedding, $i =1, ..., 34$, given a classification-threshold parameter value between $(x_9, x_{10}) = (0.0528, 0.0556)$ for the nodes 9 and 10.  

However, computational efficiency and scalability can be challenging to any of the aforementioned approaches, if the implementation requires full-graph operations. For instance, full-graph training of GCN can be hampered by slow convergence and scales poorly to very large graphs (Liu et al., 2021). 

Sampling can be beneficial to both computational efficiency and scalability. A general statistical framework  for graph sampling has been developed in the recent years (Zhang, 2021a; Zhang and Patone, 2017), including the theories for probability-based breadth-first or depth-first sampling methods. In this paper, we propose two sampling-based approaches to node embedding, which aims to extract useful features from both the eigenvectors related to the graph Laplacien and the given values associated with the graph. 

In the first approach, node embeddings are given by a parametric neighbour function that holds exactly (or approximately) for the eigenvectors of a chosen Laplacian matrix. The function value of any given node depends directly on those of its adjacent nodes and, thus, recursively on all the nodes connected to it in the graph. The parameters of the neighbour function can be defined given the whole graph and the associated values. Then, for situations where full-graph training needs to be avoided, supervised learning from sample graphs can yield valid estimates of the full-graph parameters. 

In the second approach, nonparametric supervised embeddings are first generated for the sampled nodes. These are then updated by repeated sampling from the given graph, i.e. for the nodes that are sampled on each occasion. The approach facilitates parallel computing, where many small sample graphs can be processed separately from each other following the same procedure. In the extreme case, the required graph computation can be minimised by letting each node sample consist of only one selected node and its adjacent nodes. As will be illustrated, although the expected node embedding over all possible samples would differ somewhat to that based on the full-graph training, they can be just as useful for node classification tasks.

\section{Sampling-based estimation}

Let $G = (U,A)$ be an undirected simple graph with $U$ as the set of nodes, $N = |U|$, and $A$ as the adjacency matrix, where $a_{ij} = 1$ if an edge exists from $i$ to $j$ and 0 otherwise, and $a_{ij} \equiv a_{ji}$ for any $i,j\in U$. Provided they are available, let $v_G = (y_U, \omega_A)$ contain all the given values (or vectors) associated with the nodes and edges, where $y_U = \{ y_i : i\in U\}$ and $\omega_A = \{ \omega_{ij} : i,j \in U, a_{ij} =1 \}$. 

Let node embedding yield $x_U = \{ x_i : i\in U\}$, where $x_i$ is the feature value (or vector) of node $i$. Let $\theta$ be the unknown constants required for node embedding, such as the parameters of a model of $x_U$ or simply $\theta = x_U$. Given $(G, v_G)$, let the graph-fit of $\theta$, denoted by $\theta_0$, be a solution to the estimating equation 
\[
u(\theta_0) = \frac{\partial \Delta(\theta)}{\partial \theta}|_{\theta = \theta_0} = 0
\]
where we require the loss function $\Delta(\theta)$ to be a sum over all the nodes, i.e.
\[
\Delta(\theta) = \sum_{i\in U} \Delta_i(\theta)
\]
Note that $\Delta_i$ may depend on the node $i$ as well as its adjacent nodes, denoted by $\nu_i = \{ j\in U : a_{ij} a_{ji} = 1\}$ and referred to as its neighbourhood in the graph, but we write simply $\Delta_i$ (instead of $\Delta_{i,\nu_i}$) insofar as $\nu_i$ is fixed given $i$.

Denote by $G_s = (U_s, A_s)$ a sample graph that is selected from $G$ according to a given graph sampling method, where $U_s$ contains the observed nodes of $U$ and $A_s$ the observed elements of the adjacency matrix $A$. In particular, the way by which the edges drive graph sampling is called the \emph{observation procedure (OP)}; see Zhang (2021a), Zhang and Patone (2017). For undirected simple graphs considered in this paper, we assume the OP is incident such that, given an initial sample of nodes $s_0$, both $a_{ij}$ and $a_{ji}$ are observed for any $i\in s_0$. Moreover, the values in $v_G$ are assumed to be observed together with the associated nodes and edges. Finally, an OP may be applied repeatedly, i.e. to the observed nodes outside the initial $s_0$, and the set of nodes to which the OP is applied is called the \emph{seed sample}, denoted by $s$. By the general definition of sample graph in Zhang (2021a), $A_s$ contains then all the elements of $A$ which are observed from the seed sample, and $U_s = s \cup \{ i,j : a_{ij}\in A_s, a_{ij} =1\}$. 

Given $G_s$, we shall obtain an estimator of $\theta_0$, denoted by $\hat{\theta}_0$, by solving a \emph{sample estimating equation (SEE)}, such as
\[
u_s(\hat{\theta}_0) = \sum_{i\in U_s} w_i u_i(\hat{\theta}_0) =0
\]
where $u_i(\hat{\theta}_0) = \partial \Delta_i/\partial \theta|_{\theta = \hat{\theta}_0}$ and $w_i$ is some chosen weight. In particular, the SEE is \emph{unbiased} over 
repeated sampling if, for any $\theta_0$, we have
\begin{equation} \label{SEE}
E\{ u_s(\theta_0) \} = 0 
\end{equation}
where the expectation is over the sampling distribution of $G_s$ given fixed $(G, v_G)$.

\subsection{Targeted random walk (TRW)}  \label{TRW}

There are many kinds of random walk in graphs; see e.g. Masuda et al. (2017). We consider the TRW (Zhang, 2021a).
Starting from any node, $O_0 \in U$, the transition probabilities from $O_t = i$ to $O_{t+1} = j$ are given by
\[
p_{ij} = \begin{cases} \frac{1}{d_i + r} \big( 1 + \frac{r}{N} \big) & \text{if } a_{ij} =1 \\ 
\frac{r}{d_i +r} \big( \frac{1}{N} \big) & \text{if } a_{ij} = 0 \end{cases}
\]
where $d_i$ is the node degree and $r$ is a chosen tuning constant (Avrachenkov, et al., 2010). That is, one either moves randomly over one of the edges incident to the current node or jumps randomly to any node (including the current one). The Markov process $\{ O_t : t\geq 0 \}$ is irreducible by virtue of random jump. At equilibrium, the stationary probability is given by
\[
\pi_i = \Pr(O_t = i) \propto d_i + r
\]
The same $\pi_i$ can as well be obtained by faster-mixing lagged random walk (Zhang, 2021b), or one can design any individual stationary probabilities by lagged Metropolis-Hastings walk (Zhang, 2022).

Applying the OP to the current $O_t = i$ enables one to observe $\nu_i$ and the associated values $y_{\nu_i} = \{ y_j : j\in \nu_i \}$ and $\omega_{\nu_i} = \{ (\omega_{ij}, \omega_{ji}) : j\in \nu_i\}$, if they exist. Given any extraction of $n$ states at equilibrium, as the seed sample of TRW and denoted by $s = \{ t_1, ..., t_n\}$, an unbiased SEE \eqref{SEE} can be given as
\[
u_s(\theta) = \sum_{k=1}^n \sum_{i\in U} \mathbb{I}(O_{t_k} =i) \frac{u_i(\theta)}{n(d_i +r)} 
\]
since
\[
E\{ u_s(\theta_0) \} = \frac{1}{n} \sum_{k=1}^n \sum_{i\in U} \frac{\pi_i}{d_i + r} u_i(\theta_0) \propto u(\theta_0) = 0
\]

Under TRW, the states $\{ O_{t_1}, ..., O_{t_n} \}$ are correlated. The simplest approach to unbiased variance estimation is to run multiple TRWs independently and obtain, say, $\hat{\theta}_{0,l}$ for $l =1, ..., L$. The combined estimator and variance estimator are then given by
\[
\hat{\theta}_0 = \frac{1}{L} \sum_{l=1}^L \hat{\theta}_{0,l} \qquad\text{and}\qquad 
\hat{V}(\hat{\theta}_0) = \frac{1}{L(L-1)} \sum_{l=1}^L (\hat{\theta}_{0,l} - \hat{\theta}_0)^2 
\]

\subsection{Snowball sampling (SBS)} \label{TSBS}

Let $s_0$ be an initial node sample, $s_0 \subset U$. For $t=1, ..., T$, let
\[
s_t = \nu(s_{t-1}) \setminus \bigcup_{r=0}^{t-1} s_r 
\]
be the $t$-th wave sample of nodes, which are outside of $s_{t-1}$ but adjacent to some nodes in $s_{t-1}$. The sampling is terminated if $s_t = \emptyset$ for some $t< T$, in which case $s_r = \emptyset$ for $r = t, ..., T$. The seed sample of $T$-wave SBS is 
\[
s =\bigcup_{t=0}^{T-1} s_t
\]
by repeated incident OP, and the elements of $A_s$ are $s\times U \cup U\times s$. See Zhang (2021a), Frank and Snijders (1994), and Goodman (1961).

Different strategies (Zhang, 2021a, Ch. 5) can be used to define the SEE. For instance, given the sample graph $G_s = (U_s, A_s)$, let $F_{i,s}$ contain all the nodes in $G_s$ that can lead to $i\in s$ by $T$-wave SBS, in contrast to $F_i$ that contains all such nodes in $G$. We can let $\delta_i =1$ only if $i\in s$ results from $s_0\cap F_{i,s} \neq \emptyset$, and 0 otherwise --- including when $i\in s$ only because of $s_0 \cap (F_i\setminus F_{i,s}) \neq \emptyset$ --- although the distinction vanishes in the special case of 1-wave SBS, where 
\[
s = s_0 \qquad\text{and}\qquad  F_{i,s} = F_i = \nu_i \cup \{ i\} 
\]
Next, let
\[
w_i^{-1} = \Pr(s_0\cap F_{i,s} \neq \emptyset) = E(\delta_i)
\]
according to the design of the initial sample $s_0$. We have then
\[
E\{ u_s(\theta_0) \} = \sum_{i\in U} E(\delta_i) w_i u_i(\theta_0) = u(\theta_0) =0 
\]
Moreover, by expansion around $\theta_0$, we have
\[
0 = \sum_{i\in s} w_i u_i(\hat{\theta}_0) \approx \sum_{i\in s} w_i u_i(\theta_0) + \mathcal{H}_s (\hat{\theta} - \theta_0)
\]
where $\mathcal{H}_s = E\big( \sum_{i\in s} w_i \partial u_i(\theta)/\partial \theta|_{\theta =\theta_0}\big)$. The sampling variance of $\hat{\theta}_0$ follows as
\[
V(\hat{\theta}_0) \approx \mathcal{H}_s^{-1} 
\left\{ \sum_{i,j\in U} \Big( w_i w_j \Pr( \delta_i \delta_j =1) - 1\Big) u_i(\theta_0) u_j(\theta_0)' \right\}
\mathcal{H}_s^{-1} 
\]
We refer to Zhang (2021a) for relevant details of T-wave SBS.

\section{Eigen neighbour function} 

Let $L = D - A$ be the Laplacian matrix of $G$, where $D$ is the diagonal matrix of degrees $d_i$. The normalised graph Laplacian is 
\[
\dot{L} = D^{-\frac{1}{2}} L D^{-\frac{1}{2}} = I - D^{-\frac{1}{2}} A D^{-\frac{1}{2}} = I - M 
\]
Denote by $\lambda_0$ the smallest non-zero eigenvalue of $\dot{L}$ and by $z_0$ the corresponding eigenvector. The components of $z_0/\sqrt{d}$ are more similar for adjacent nodes than any other eigenvector $z$ with corresponding non-zero eigenvalue $\lambda_z$. Thus, $z_0$ may be a useful embedding for any node values that are more similar among adjacent nodes than otherwise, an example of which is the ZKC members split. However, one generally cannot estimate $z_0$ based on a subgraph of $G$. 

Regarding any binary node classification $y$, for which the eigenvector $z_0$ is a good predictor, consider generalised linear node classification in combination with eigen neighbour function (ENF) for embedding. For any $i\in U$, let 
\begin{equation} \label{GL-ENF}
\begin{cases} \eta_i = g(p_i) = (1, x_i) \psi \\ x_i = \xi \sum_{j\in \nu_i} m_{ij} y_j \end{cases}
\end{equation}
where $g(\cdot)$ is a link function mapping $p_i = p(x_i; \psi) \in (0,1)$ to the linear scale, $x_i$ is a scalar feature (or embedding) of node $i$ which depends on its adjacent nodes, $m_{ij}$ is an element of the normalised adjacency matrix $M$, $\xi$ is a scalar parameter for embedding and $\psi$ is a (column) vector parameter for classification given $x_i$. Both $\xi$ and $\psi$ are parameters that need to be estimated.

\paragraph{Classification} One may consider $p_i$ as a probability for binary $y_i$, according to which a classifier can simply be  $\mathbb{I}(p_i > 0.5)$. Given $(G, x_U)$, the graph-fit $\psi_0$ can be obtained by the loss and estimation equation below
\begin{align*}
& \Delta(\psi) \propto \sum_{i\in U} y_i \log(p_i) + (1- y_i) \log(1 - p_i) \\
& u(\psi) = \frac{\partial \Delta(\psi)}{\partial \psi} = \sum_{i\in U} \big( \frac{\partial p_i}{\partial \psi} \big) \frac{y_i - p_i}{p_i (1-p_i)}  =0 
\end{align*}
For example, the logistic link yields $p_i = 1/\big( 1 + e^{-\eta_i} \big)$, such that
\[
u(\psi) = \sum_{i\in U} (y_i - p_i) (1, x_i)'  \quad\text{and}\quad
\frac{\partial u(\psi)}{\partial \psi} = - \sum_{i\in U} p_i (1- p_i) (1, x_i)' (1, x_i) 
\]
The tanh link yields $p_i = \frac{1}{2} (1+t_i)$ where $t_i = \big( e^{\eta_i} - e^{-\eta_i} \big)/\big( e^{\eta_i} + e^{-\eta_i} \big)$, such that
\[
u(\psi) = \sum_{i\in U} (y_i - p_i) (1, x_i)'  \quad\text{and}\quad
\frac{\partial u(\psi)}{\partial \psi} = - \frac{1}{2} \sum_{i\in U} (1+t_i) (1- t_i) (1, x_i)' (1, x_i) 
\]

\paragraph{Embedding} The underlying neighbour function for $x_i$ in \eqref{GL-ENF} can be written as 
\[
\mu = \xi \dot{\mu} \qquad\text{and}\qquad \dot{\mu} = M \mu
\]
in terms of an $N$-vector $\mu' = (\mu_1, ..., \mu_N)$. Since any eigenvector $z$ of $\dot{L}$ satisfies $z = (1- \lambda_z)^{-1} M z$, the feature vector $x = \xi M y$ would hopefully combine useful information in the eigenvectors of $\dot{L}$ and the given labels $y$; hence  the term ENF. Let $e = y - \mu$ be the discrepancy of $\mu$ satisfying $\mu = \xi_0 \dot{\mu}$, such that
\[
y - \xi_0 \dot{y} = (\mu + e) - \xi_0 (\dot{\mu} + \dot{e}) = e - \xi_0 \dot{e}
\]
Let the total loss be 
\begin{equation} \label{ENF-loss}
\Delta(\xi) = \sum_{i\in U} \Delta_i(\xi) = \sum_{i\in U} (y_i - \xi \dot{y}_i)^2 
\end{equation}
Given $(G, y_U)$, the graph-fit $\xi_0$ minimising $\Delta(\xi)$ is given by
\[
\xi_0 = \big( \dot{y}' \dot{y} \big)^{-1} \big( \dot{y}' y \big)
\]

\paragraph{Supervised learning} Estimates of the graph-fit $(\xi_0, \psi_0)$ can be obtained from the respective SEEs given any sample graph $G_s$ and associated $y$-values. In particular, under 1-wave SBS with seed sample $s=s_0$, yielding $U_s = \bigcup_{i\in s} \nu_i \cup s$ and the associated $y_{U_s}$, let the SEE for $\xi$ be
\[
u_s(\xi) = \frac{\partial}{\partial \xi} \sum_{i\in s} \pi_i^{-1} (y_i - \xi \dot{y}_i)^2 
\propto \sum_{i\in s} \pi_i^{-1} \dot{y}_i (y_i - \xi \dot{y}_i) =0
\] 
such that
\[
\hat{\xi}_0 = \Big( \sum_{i\in s} \frac{\dot{y}_i^2}{\pi_i} \Big)^{-1} \Big( \sum_{i\in s} \frac{\dot{y}_i y_i}{\pi_i} \Big)
\]
Moreover, let $\hat{x}_i = \hat{\xi}_0 \dot{y}_i$ and obtain $\hat{\psi}_0$ from the plug-in SEE for $\psi$ given by
\[
u_s(\psi) = \sum_{i\in s} \pi_i^{-1} (y_i - p_i) (1, \hat{x}_i)' = 0
\]

\subsection{Example: ZKC}

Denote the members split shown in Figure \ref{fig:ZKC} by a binary label $y_i$, for all $i\in U$. As mentioned before, given the eigenvector $z_0$, a perfect classifier is $y_i = \mathbb{I}(z_{0,i} > \tau)$ for any $i\in U$, given any parameter value $\tau \in (0.0528, 0.0556)$. 

\begin{figure}[ht]
\centering
\includegraphics[width=140mm, height=65mm]{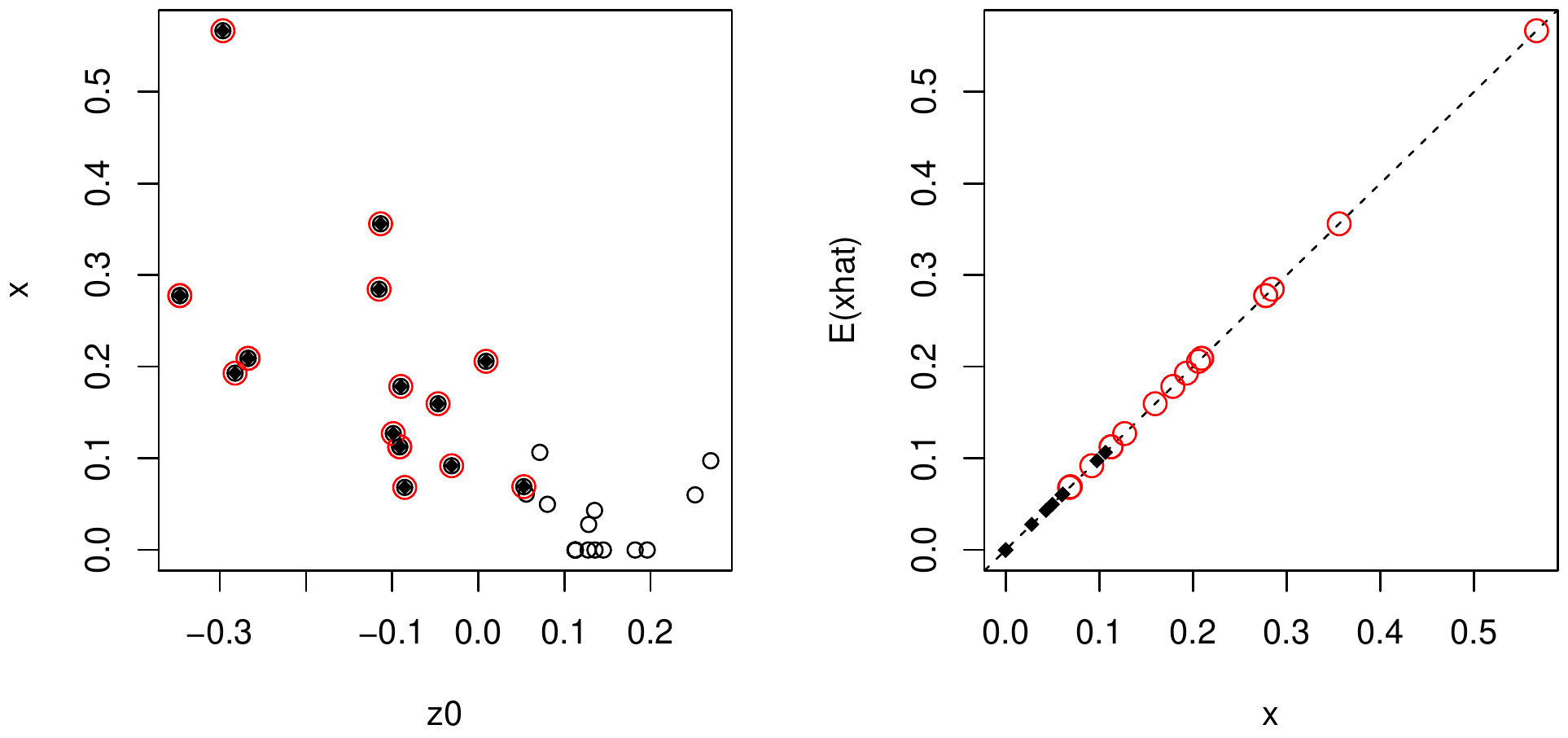} \\ \bigskip
\includegraphics[width=140mm, height=65mm]{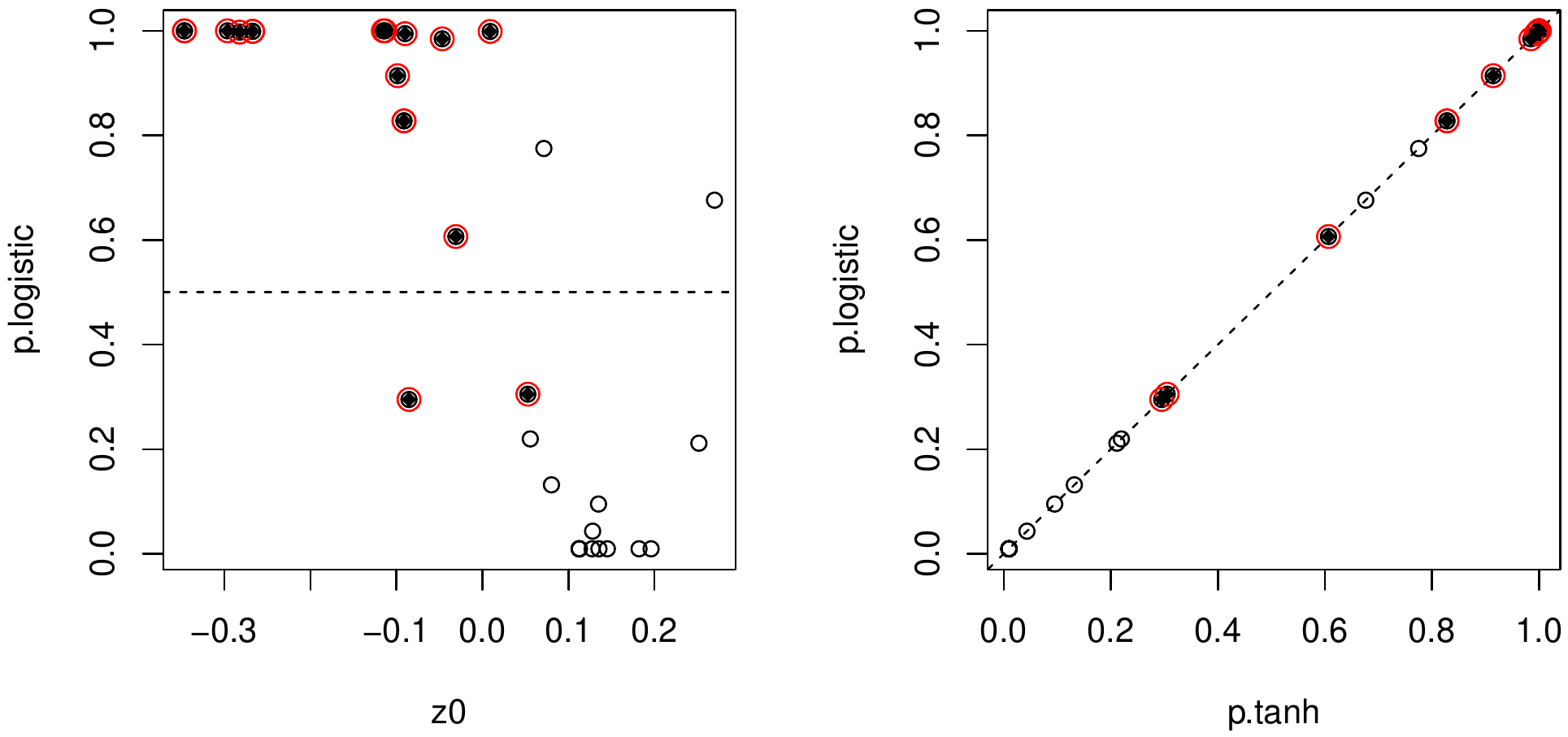}
\caption{Normalised $x = \xi_0 \dot{y}$ against eigenvector $z_0$ (top-left) or $E(\hat{x})$ (top-right), given $\hat{x} = \hat{\xi}_0 \dot{y}$ and $\hat{\xi}_0$ by snowball sampling with $|s| =5$. Logistic-ENF $p(x; \psi_0)$ against $z_0$ (bottom-left) or tanh-ENF $p$ (bottom-right). Members split marked.} \label{fig:Z-ENF}
\end{figure}

The top-left plot of Figure \ref{fig:Z-ENF} shows the normalised ENF embedding $x = \xi_0 \dot{y}$ given the graph-fit $\xi_0 = 0.955$, where $x'x=1$. The ENF yields a feature vector $x$ that is similar to $z_0$, but less discriminant than $z_0$ for the $y$-classification, since there clearly exists an interval of $x$ in which $y$ can be both 0 and 1. The top-right plot of Figure \ref{fig:Z-ENF} shows $E(\hat{x})$ against $x$, where $\hat{x} = \hat{\xi}_0 \dot{y}$ and $\hat{\xi}_0$ is obtained by 1-wave SBS with initial simple random sampling (SRS) of sample size $|s| =5$. The sample-graph estimator $\hat{\xi}_0$ is unbiased for $\xi_0$ and embedding $x$.

The bottom-left plot of Figure \ref{fig:Z-ENF} shows the logistic-ENF $p(\xi_0 \dot{y}; \psi_0)$ against the eigenvector $z_0$ given the graph-fit $\psi_0 = (-4.631, 15.747)$. Applying $\mathbb{I}(p_i > 0.5)$ to all the 34 nodes would incorrectly classify two 1-nodes and two 0-nodes. The bottom-right plot of Figure \ref{fig:Z-ENF} shows the logistic-ENF against the tanh-ENF, given the graph-fit $\psi_0 = (-2.315, 7.874)$ for the latter, which shows that the choice of link does not matter for this valued graph.

\subsection{Discussion}

In the statistical literature of spatial and network analysis, e.g. Ord (1975), Friedkin (1990) and Leenders (2002), the linear autoregressive predictor
\[
\mu = \rho H \mu + c  \qquad\Leftrightarrow\qquad \mu = (I - \rho H)^{-1} c
\]
builds on node features $c_i = y_i' \beta$ for $i\in U$, where $\beta$ is a parameter vector, $H = [h_{ij}]$ and $h_{ij} = 0$ for any $j\notin \nu_i$. Provided $\| \rho H\| <1$, we have 
\[
\mu = (I + \rho H + \rho^2 H^2 + \cdots) c
\]
showing how $\mu_i$ depends on $c_i$, those of its 1-hop neighbours, 2-hop neighbours, etc. For $l=1, ..., L$, the $l$-th order approximation is given by iterating
\[
\mu^{(l)} = \rho H \mu^{(l-1)} + c \quad\text{and}\quad \mu^{(0)} = c
\]
Such network influencing has certain similarity to messaging-and-aggregating by an $L$-layer GNN (e.g. Hamilton, 2020), e.g.
\[
\mu^{(l)} = \sigma\Big( \rho_l H \mu^{(l-1)} + \phi_l \mu^{(l-1)} \Big)
\]
where $\sigma(\cdot)$ is an activation function such as $\max(0,\cdot)$, and $\{ (\rho_l, \phi_l) : 1\leq l \leq L\}$ are the parameters to be learned. The initial $\mu^{(0)} = c$ can be directly given by known node features or functions of them. 

The ENF $\dot{x} = My$ is a one-layer GNN with identity activation function $\sigma(\cdot)$, where $x^{(0)} = y$, $H = M$, $\xi = \rho_1$ and $\phi_1=0$. The sampling-based training approach to ENF above can be adapted to other model-based embeddings. 
For instance, let $x^{(0)} = y$, and let $x$ be given by a 2-layer GNN as
\[
x = \sigma\Big( \rho_2 H \sigma\big(\rho_1 H y + \phi_1 y\big) + \phi_2 \sigma\big(\rho_1 H y + \phi_1 y\big) \Big) 
= \mathring{y}
\]
with parameters $\xi = (\rho_2, \rho_2, \phi_2, \phi_1)$. Similarly to \eqref{ENF-loss}, one can use the loss  
\[
\Delta_i(\xi) = (y_i - \mathring{y}_i)^2
\]
for the underlying 2-hop neighbour function satisfying $\mu_i = \mathring{\mu}_i(\xi_0)$. For sampling-based estimation of $\xi_0$, one now needs to observe $y_j$ of any node $j$ that belongs to the 2-hop neighbourhood of a seed sample node $i$. In the case of $s =s_0$, one can implement the 2-hop incident OP, which yields $A_s = \tilde{s}\times U \cup U \times \tilde{s}$ where $\tilde{s} = \bigcup_{i\in s} \nu_i \cup s$, and $U_s = \bigcup_{i\in s} \nu_i^2 \cup s$ where $\nu_i^2 = \bigcup_{j\in \nu_i} \nu_j$, and the associated $y_{U_s}$.

\section{Normalised Laplacian embedding} 

Consider a different approach to eigenvector-like embeddings, which operates without an explicit model but through an appropriately chosen linear system $P z = 0$, which is satisfied exactly (or nearly so) by any eigenvector $z$ of $\dot{L}$. In particular, for any pair of eigenvalue and eigenvector $(\lambda_z, z)$, we can let
\[
P_{\lambda_z}(M) =  (1- \lambda_z) I - M
\]
or
\[
P_{\lambda_z}(\tilde{M}) = \mbox{Diag}\Big( 1 - \frac{\lambda_z d_i}{1 + d_i} \Big) - \tilde{M}
\]
where 
\[
\tilde{M} = (I +D)^{-\frac{1}{2}} (I+A) (I+D)^{-\frac{1}{2}} 
\]
It is convenient to define $\tilde{\nu}_i = \{ j\in U : p_{ij} \neq 0\}$ as the looped neighbourhood of node $i$, to be distinguished from $\nu_i$, since $P$ has non-zero diagonal elements. Note that both $P_{\lambda}(M)$ and $P_{\lambda}(\tilde{M})$ are symmetric matrices.

Given any tuning constant $\lambda \in (0,2)$, let the corresponding ($y$-) supervised normalised Laplacian embedding (SNLE) be given by minimising
\begin{equation} \label{SL-NLE}  
\Delta(x) = x' P_{\lambda}' P_{\lambda} x + \gamma (y - x)' (y - x)  
\end{equation}
over $x$, where $P_{\lambda}$ is given by replacing $\lambda_z$ with $\lambda$ in $P_{\lambda_z}$, and $\gamma$ is a regularisation constant. Setting $\partial \Delta(x)/\partial x$ to $0$ yields the graph-fit 
\[
x_0 = \big( I + \gamma^{-1} P_{\lambda}' P_{\lambda} \big)^{-1} y 
\]
Any $x$ satisfying $x' P_{\lambda}' P_{\lambda} x \approx 0$ would entail $P_{\lambda} x \approx \bm{0}$, analogously to $P_{\lambda_z} z = 0$. As we illustrate below, although the loss\eqref{SL-NLE} is defined whether or not $\lambda$ is an eigenvalue of $\dot{L}$, it can generate embeddings that are correlated with any eigenvector of the normalised Laplacian of the graph --- hence the term SNLE.

\subsection{Example: ZKC}

\begin{figure}[ht]
\centering
\includegraphics[width=140mm, height=65mm]{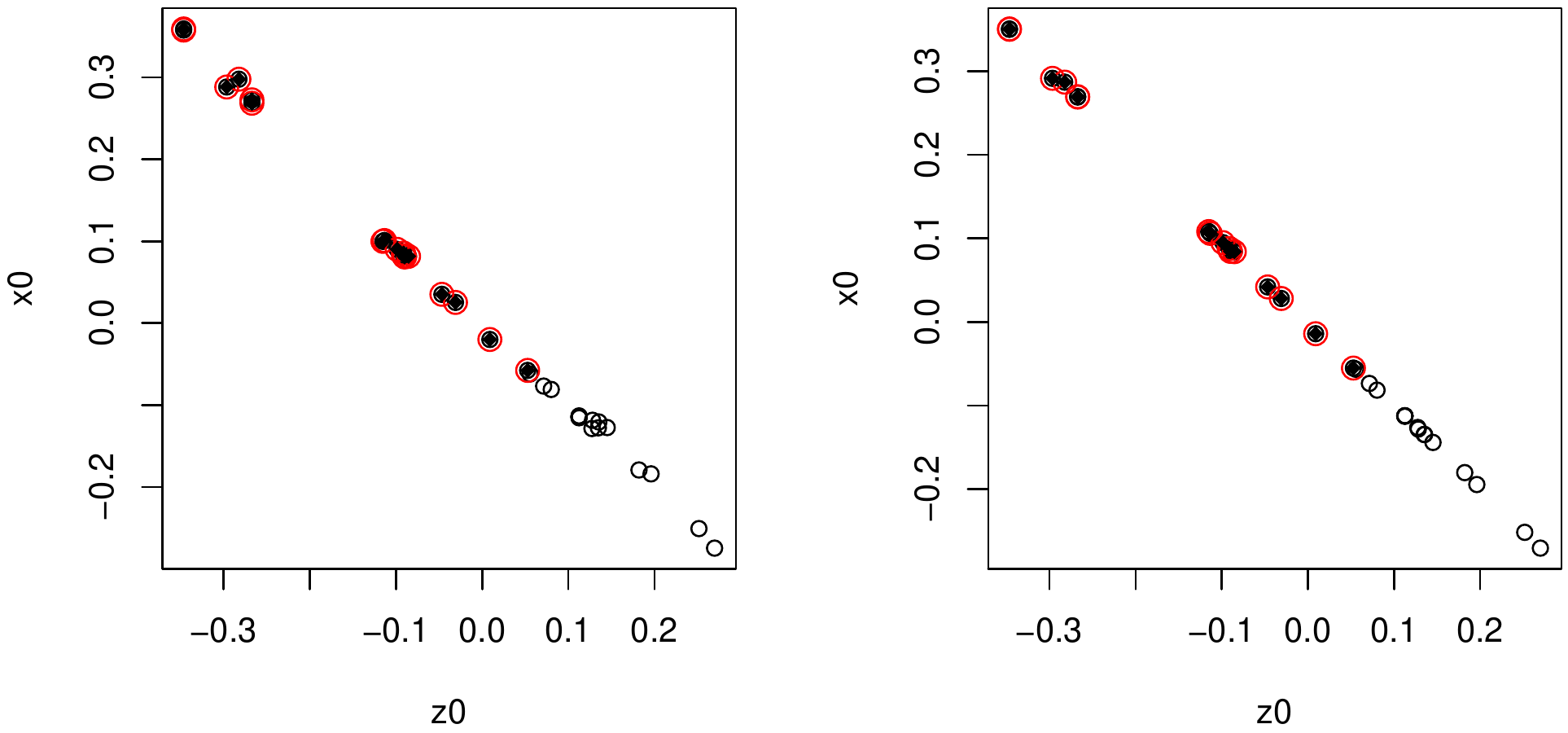}
\caption{Eigenvector $z_0$ vs. normalised $x_0$ given $\lambda = 0.1$ (left) or $0.15$ (right)} \label{fig:Z-NLE}
\end{figure}

We have $\lambda_0 = 0.132$ for the ZKC graph (Figure \ref{fig:ZKC}). As illustrated in Figure \ref{fig:Z-NLE}, full-graph embedding $x_0$ can closely resemble $z_0$ for a range of $\lambda$, given $\gamma= 0$. 
 
\begin{figure}[ht]
\centering
\includegraphics[width=140mm, height=65mm]{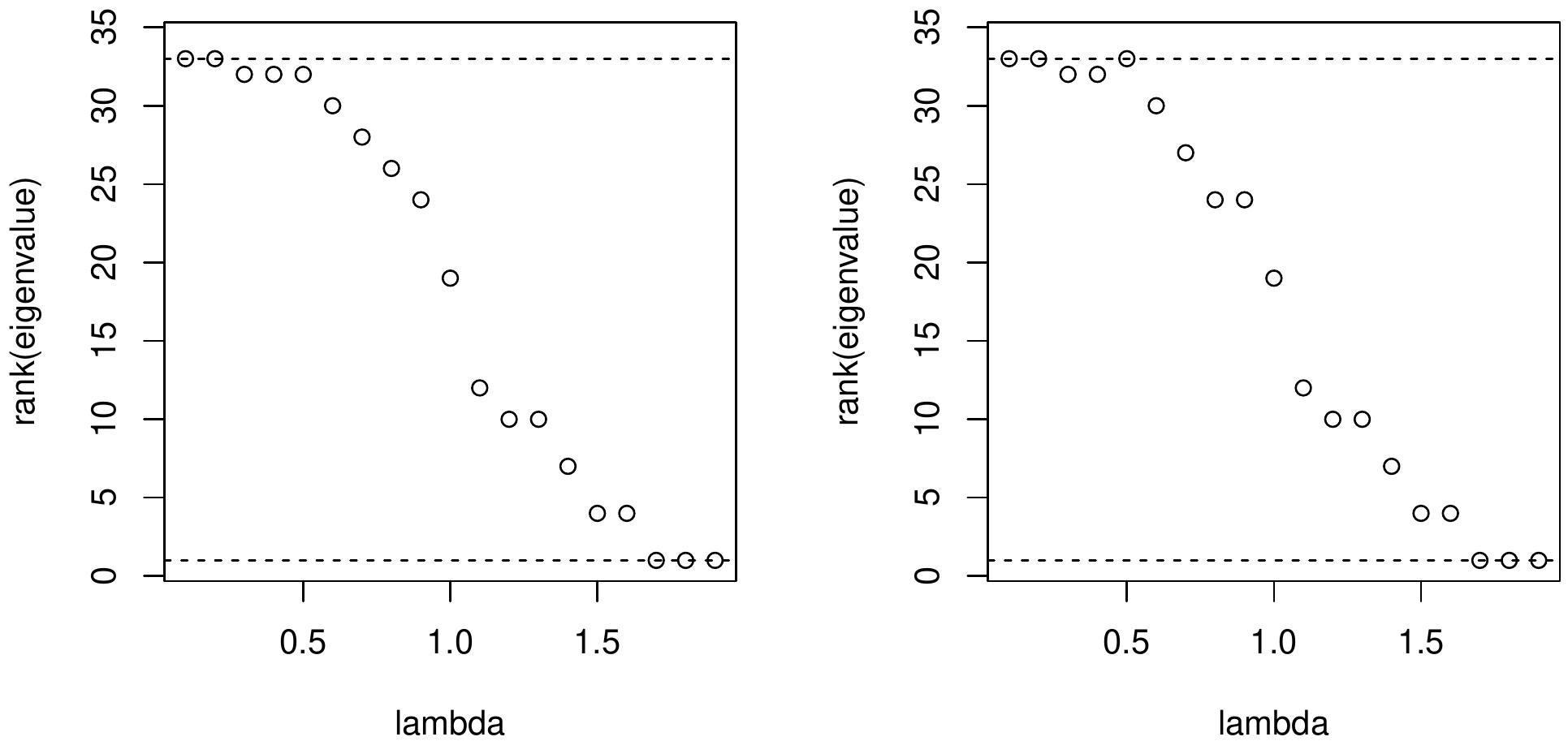}
\caption{Ranks of best-correlated eigenvector with SNLE $x_0$ as $\lambda$ varies, with fixed $\gamma \equiv 0.001$. Left, $P_{\lambda}(M)$; right, $P_{\lambda}(\tilde{M})$.} \label{fig:SL-NLE}
\end{figure}

Figure \ref{fig:SL-NLE} shows the rank of the eigenvector that is best correlated with $x_0$ given by \eqref{SL-NLE}, as $\lambda$ varies from 0.1 to 1.9 while $\gamma$ is fixed at 0.0001. The eigenvectors are ranked from 1 to 34 in the decreasing order of their corresponding eigenvalues, where the rank of $z_0$ is 33 (marked top horizontal dashed line). It is seen that SNLE \eqref{SL-NLE} can yield different eigenvector-like embeddings as $\lambda$ varies. Note that the non-monotonicity seen here for $P_{\lambda}(\tilde{M})$ can be removed by reducing $\gamma$, e.g. to 0.0001, the details of which are omitted here.

\subsection{Sample-graph embedding}

Rewrite the loss $\Delta(x)$ in \eqref{SL-NLE} as a sum over all the nodes as
\[
\Delta(x) = \sum_{i\in U} \dot{x}_i^2 + \gamma \sum_{i\in U} (y_i - x_i)^2 
\]
where $\dot{x}_i = \sum_{j\in \tilde{\nu}_i} p_{ij} x_j$. By 1-wave SBS with seed sample $s =s_0$, one observes $U_s = \bigcup_{i\in s} \nu_i \cup s = \bigcup_{k\in s} \tilde{\nu}_k$ and the associated $y_{U_s}$, as well as the sub-matrices $A_s = A_{s U_s} \cup A_{U_s s}$, where $A_{sU_s} = [a_{kj}]_{n\times n_{U_s}}$ for $k\in s$ and $j\in U_s$ and $n = |s|$ and $n_{U_s} = |U_s|$. Given any fixed $N$-vector $x$, an unbiased estimator of $\Delta(x)$ over repeated sampling of $G_s = (U_s, A_s)$ is given by
\[
\Delta_s 
= \sum_{k\in s} \pi_k^{-1} \dot{x}_k^2 + \gamma \sum_{j\in U_s} \dot{\pi}_j^{-1} (y_j - x_j)^2
\]
where $\pi_k = \Pr(i\in s)$, and $\dot{\pi}_j = \Pr(j\in U_s)$. For instance, under SRS of $s$, we have $\pi_k = n/N$ and $\dot{\pi}_j = 1 - \binom{N -d_j -1}{n}/\binom{N}{n}$.

Given any $(G_s, y_{U_s})$, supervised learning is feasible for $j\in U_s$. We have
\[
\frac{1}{2} \frac{\partial \Delta_s}{\partial x_j} = \sum_{k\in s\cap \tilde{\nu}_j} \frac{p_{jk}}{\pi_k} \sum_{l\in \tilde{\nu}_k} p_{kl} x_l 
- \frac{\gamma}{\dot{\pi}} (y_j - x_j) 
\]
where $\partial \dot{x}_k/\partial x_j = p_{jk}$ if $k\in s\cap \tilde{\nu}_j$ given $k\in s$, and 0 otherwise. 
Let $y_{U_s}$ and $x_{U_s}$ be vectors associated with $U_s$. Let $P_{sU_s} = [p_{kj}]_{n\times n_{U_s}}$ for $k\in s$ and $j\in U_s$. 
Define the $n\times n$-matrix $W_s = \mbox{Diag}(\pi_i^{-1})$ associated with $s$, and the $n_{U_s} \times n_{U_s}$-matrix $W_{U_s} = \mbox{Diag}(\dot{\pi}_j^{-1})$ associated with $U_s$. Let the SEE be given by
\[
u_s = \frac{\partial \Delta_s}{\partial x_{U_s}} = P_{sU_s}' W_s P_{sU_s} x_{U_s} - \gamma W_{U_s} (y_{U_s} - x_{U_s}) = 0
\]
Let the solution to the SEE be
\[
\hat{x}_{U_s} = \big( \gamma^{-1} W_{U_s}^{-1} P_{sU_s}' W_s P_{sU_s} + I \big)^{-1} y_{U_s}
\]
For any $i\in U$, let the sample-graph SNLE $x_i$ over repeated sampling to be
\begin{equation} \label{sNLE}
x_i = E(\hat{x}_i \mid i\in U_s)
\end{equation}

Whereas the sample-graph embedding \eqref{sNLE} surely converges to the full-graph embedding \eqref{SL-NLE} as $|s| \rightarrow |U|$, the two are unequal generally, unless $\tilde{\nu}_j \equiv s\cap \tilde{\nu}_j$ for any $j\in U_s$ under graph sampling. The reason is that $\partial \Delta/\partial x_j$ when $\nu_j$ is fully observed may differ to $\partial \Delta_s/\partial x_j$, for any $j\in U_s\setminus s$, i.e.
\[
\frac{\partial \Delta}{\partial x_j} = 2 \sum_{k\in \tilde{\nu}_{j}} p_{jk} \dot{x}_k \neq \frac{\partial \Delta_s}{\partial x_j} 
= 2 \sum_{k\in s\cap \tilde{\nu}_j} p_{jk} \dot{x}_k 
\]
Thus, derivation and expectation are not exchangeable,
\[
E(\partial \Delta_s/\partial x_j) \neq \partial E(\Delta_s)/\partial x_j
\]
The matter is the same for any graph sampling method, as long as there is a distinction between a seed sample node (whose neighbourhood is observed by definition of incident OP) and another sample node in $U_s$ (whose neighbourhood is not always observed when sampling is terminated).

\begin{figure}[ht]
\centering
\includegraphics[width=140mm, height=65mm]{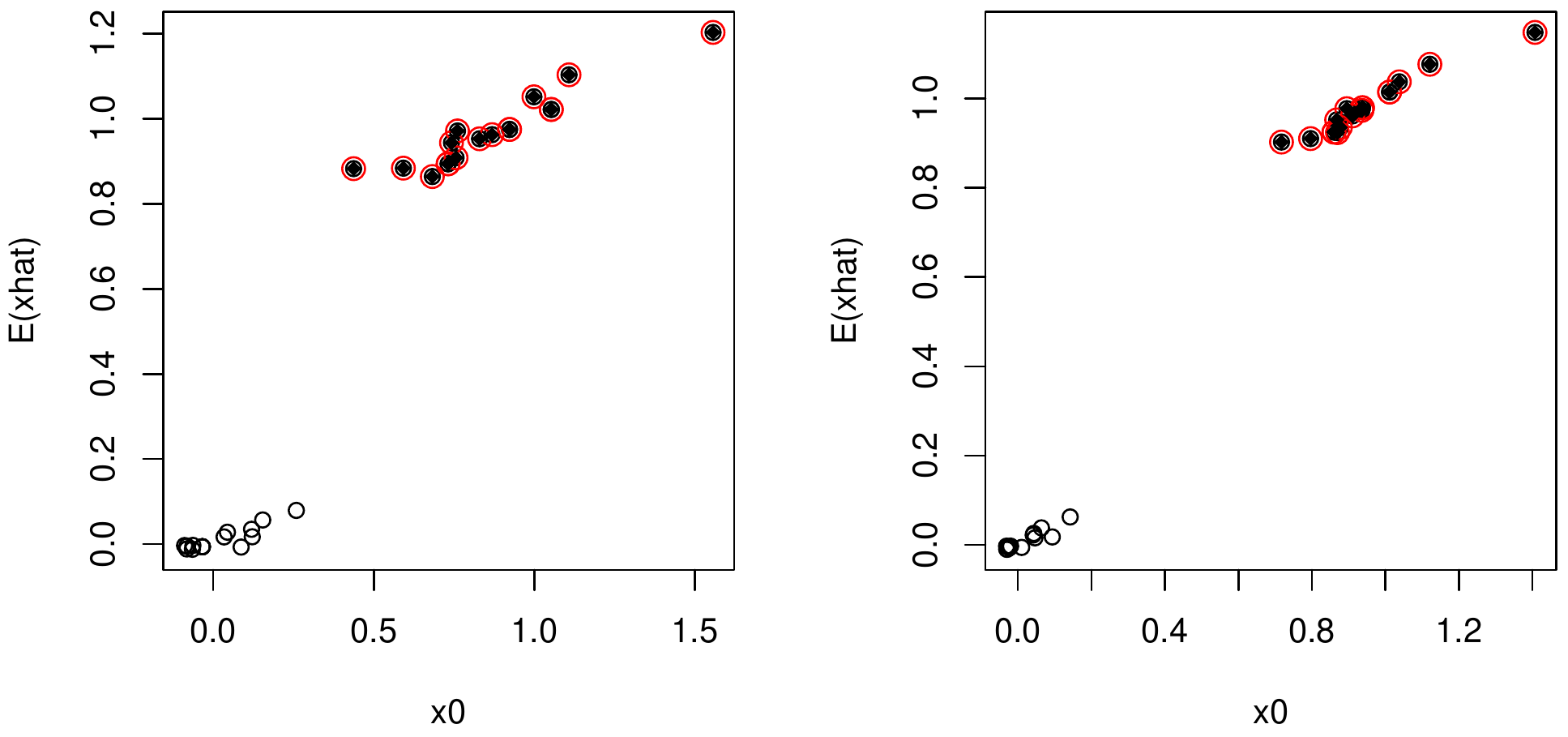}
\caption{SNLE $x_0(G,y_U)$ vs. sample embedding $E(\hat{x})$ given $\gamma=0.1$ (left) or $\gamma=1$ (right), using $\lambda=0.1$, $P_{\lambda}(\tilde{M})$, seed sample size $|s|=1$. Members split marked.} \label{fig:NLEs}
\end{figure}

Nevertheless, Figure \ref{fig:NLEs} compares full-graph embedding \eqref{SL-NLE} for the ZKC data to sample-graph embedding by \eqref{sNLE}, which is obtained by 1-wave SBS where the initial sample consists of a single randomly selected node. Both embeddings are tuned towards $z_0$ here given $P_{0.1}(\tilde{M})$. Using $|s|=1$ that minimises the required graph computation, we obtain the sample-graph embedding $E(\hat{x})$ with $\gamma = 0.1$ (left) or $\gamma =1$ (right), which readily yields $p(E(\hat{x}); \psi)$ that can be more discriminating than $p(x_0; \psi)$ or $p(z_0; \psi)$ for $y$-like labels, now that $E(\hat{x})$ separates the 1-nodes and 0-nodes by a wider margin in both plots.

\setstretch{0.9}

\end{document}